\documentclass{article}

\PassOptionsToPackage{numbers, compress}{natbib}
\usepackage{natbib}

\usepackage[preprint]{nips_2018}

\usepackage[pdftex, pagebackref, pdfpagelabels, hypertexnames, 
plainpages=false, naturalnames=false,
bookmarks, bookmarksnumbered, pdfpagemode=UseOutlines,]{hyperref}
\hypersetup{colorlinks, citecolor=[rgb]{0.0,00,0.5}, filecolor=[rgb]{0.6,0.0,0.0}, linkcolor=[rgb]{0.0,0.0,0.5}, urlcolor=[rgb]{0.0,0.0,0.5}}

\usepackage{graphicx} \graphicspath{ {figures/} } 
\usepackage{subcaption}
\usepackage{tabularx}
\newcolumntype{?}{!{\vrule width 1pt}}
\usepackage{amsmath,amsfonts}
\usepackage{algorithm}
\usepackage{algorithmic}
\usepackage{microtype}      

\title{Multi-Adversarial Variational Autoencoder Networks}

\author{
Abdullah-Al-Zubaer Imran \\
\texttt{aimran@cs.ucla.edu}
\And
Demetri Terzopoulos \\
\texttt{dt@cs.ucla.edu}
\AND
\\[-8mm] Computer Science Department\\ University of California, Los Angeles
}

\begin{document}

\maketitle

\begin{abstract}
The unsupervised training of GANs and VAEs has enabled them to generate realistic images mimicking real-world distributions and perform image-based unsupervised clustering or semi-supervised classification. Combining the power of these two generative models, we introduce Multi-Adversarial Variational autoEncoder Networks (MAVENs), a novel network architecture that incorporates an ensemble of discriminators in a VAE-GAN network, with simultaneous adversarial learning and variational inference. We apply MAVENs to the generation of synthetic images and propose a new distribution measure to quantify the quality of the generated images. Our experimental results using datasets from the computer vision and medical imaging domains---Street View House Numbers, CIFAR-10, and Chest X-Ray datasets---demonstrate competitive performance against state-of-the-art semi-supervised models both in image generation and classification tasks.
\end{abstract}

\section{Introduction}

Training deep neural networks usually requires a large pool of labeled data, yet obtaining large datasets for tasks such as image classification remains a fundamental challenge. Although there has been explosive progress in the production of vast quantities of high resolution images, large collections of labeled data required for supervised learning remain scarce. Especially in domains such as medical imaging, datasets are limited in size due to privacy issues, and manual annotation by medical experts is expensive, time-consuming, and prone to subjectivity, human error, and variance across different experts. Even when large labeled datasets become available, they are often highly imbalanced and nonuniformly distributed. For instance, in an imbalanced medical dataset there will be an over-representation of common medical problems and an under-representation of rare conditions. Such biases make the training of neural networks across multiple classes with similar effectiveness very challenging. 

The small-training-data problem is traditionally mitigated through simplistic and cumbersome data augmentation, often by creating new training examples through translation, rotation, flipping, etc. The missing or mismatched label problem can be addressed by evaluating similarity measures over the training examples. This is not always robust and the efficiency largely depends on the performance of the similarity measuring algorithms. 

\begin{figure}
\centering
\subcaptionbox{}{\includegraphics[width=0.3\linewidth,height=0.3\linewidth]{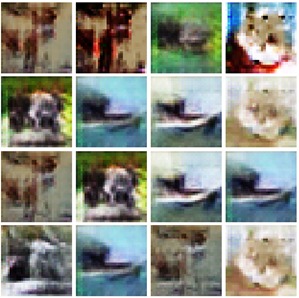}} \hfill
\subcaptionbox{}{\includegraphics[width=0.3\linewidth,height=0.3\linewidth]{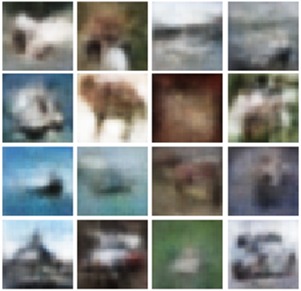}} \hfill
\subcaptionbox{}{\includegraphics[width=0.3\linewidth,height=0.3\linewidth,trim={0 0 0 1.5},clip]{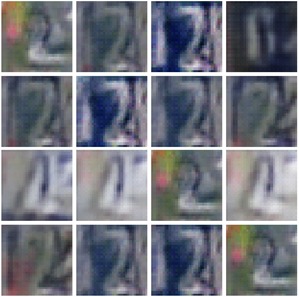}}
\caption{Synthetic images generated from CIFAR-10: (a) Relatively good generation by a GAN. (b) Blurry images generated by a VAE. From SVHN: (c) mode collapsed generation by a GAN.} 
\label{fig:gan_collapse}
\end{figure}

Generative models, such as VAEs \citep{kingma2013auto} and GANs \citep{goodfellow2014generative}, have recently become popular because of their ability to learn underlying data distributions from training samples. This has made generative models more practical in ever-frequent scenarios where there is an abundance of unlabeled data. With minimal annotation, an efficient semi-supervised learning model could be a go-to approach. More specifically, based on small quantities of annotation, generative models could be utilized to learn real-data distributions and synthesize realistic new training images. Both VAEs and GANs can be employed for this purpose. 

VAEs can learn the dimensionality-reduced representation of training data and, with an explicit density estimation, can generate new samples. However VAE-generated samples are usually blurry (Fig.~\ref{fig:gan_collapse}b). On the other hand, despite the successes in generating images and semi-supervised classifications, GAN frameworks are still very difficult to train and there are challenges in using GAN models, such as non-convergence due to unstable training, mode collapsed image generation (Fig.~\ref{fig:gan_collapse}c), diminished gradient, overfitting, and high sensitivity to hyper-parameters.

To stabilize GAN training and combat mode collapse, several variants have been proposed. \citet{nguyen2017dual} proposed a model, where a single generator is used alongside dual discriminators. \citet{durugkar2016generative} proposed a model with a single generator and feedback aggregated over several discriminators considering either the average loss of all discriminators or by picking only the discriminator with the maximum loss in relation to the generator's output. \citet{neyshabur2017stabilizing} proposed a framework where a single generator simultaneously trains against an array of discriminators, each of which operates on a different low-dimensional projection of the data. \citet{mordido2018dropout}, arguing that all the previous approaches restrict the discriminator's architecture, which compromises the extensibility of the framework, instead proposed a Dropout-GAN, where a single generator is trained against a dynamically changing ensemble of discriminators. However, there could be a risk of dropping out all the discriminators. Feature matching and minibatch discrimination techniques have been proposed \citep{salimans2016improvedT} for eliminating mode collapsing and preventing overfitting in GAN training.

Although there have been wide ranging efforts in high quality image generation with GANs and VAEs, accuracy and image quality are usually not ensured in the same model, especially in multi-class image classification. To tackle this issue, we propose a novel method that can learn joint image generation and multi-class image classification. Our specific contribution is the Multi-Adversarial Variational autoEncoder Network, or MAVEN, a novel multi-class image classification model incorporating an ensemble of discriminators in a combined VAE-GAN network. An ensemble layer combines the feedback from multiple discriminators at the end of each batch. With the inclusion of ensemble learning at the end of a VAE-GAN, both generated image quality and classification accuracy are improved simultaneously. We also introduce a simplified version of the Descriptive Distribution Distance (DDD) measure for evaluating any generative model, which better represents the distribution of the generated data and quantifies its closeness to the real data. Our experimental results on a number of different datasets in both the computer vision and medical imaging domains indicate that our MAVEN model improves upon the joint image generation and classification performance of a GAN and a VAE-GAN with the same set of hyper-parameters.

\section{Related Work}

Generative modeling has attracted much attention in the computer vision and medical imaging research communities. In particular, realistic image generation greatly helps address many problems involving the scarcity of labeled data. GANs and their variants have been applied in different architectures in continuing efforts to improve the accuracy and effectiveness of image classification. The GAN framework has been utilized in numerous works as a more generic approach to generating realistic training images that synthetically augment datasets in order to combat overfitting; e.g., for synthetic data augmentation in liver lesions \citep{frid2018gan}, retinal fundi \citep{guibas2017synthetic}, histopathology \citep{hou2017unsupervised}, and chest X-rays \citep{salehinejad2018generalization}. \citet{calimeri2017biomedical} employed a LAPGAN \citep{Denton2015DeepGI} and \citet{han2018gan} used a WGAN \citep{arjovsky2017wasserstein} to generate synthetic brain MR images. \citet{bermudez2018learning} used a DCGAN \citep{radford2015unsupervised} to generate 2D brain MR images followed by an autoencoder for image denoising. \citet{chuquicusma2018fool} utilized a DCGAN to generate lung nodules and then conducted a Turing test to evaluate the quality of the generated samples. GAN frameworks were also shown to improve accuracy of image classification via generation of new synthetic training images. \citet{frid2018gan} used a DCGAN and a ACGAN \citep{Odena2017Conditional} to generate images of three liver lesion classes to synthetically augment the limited dataset and improve the performance of CNN for liver lesion classification. Similarly, \citet{salehinejad2018generalization} employed a DCGAN to artificially simulate pathology across five classes of chest X-rays in order to augment the original imbalanced dataset and improve the performance of a CNN model in chest pathology classification. 

The GAN framework has also been utilized in semi-supervised learning architectures to help leverage the vast number of unlabeled data alongside limited labeled data. The following efforts demonstrate how incorporating unlabeled data in the GAN framework has led to significant improvements in the accuracy of image-level classification: \citet{madani2018semi} used an order of magnitude less labeled data with a DCGAN in semi-supervised learning and showed comparable performance to a traditional supervised CNN classifier. Furthermore, their study also demonstrated reduced domain over-fitting by simply supplying unlabeled test domain images. \citet{springenberg2015unsupervised} combined a WGAN and CatGAN \citep{wang2017catgan} for unsupervised and semi-supervised learning of feature representation of dermoscopy images.

Despite these successes, GAN frameworks are very difficult to train, as was discussed in the previous section. Our work mitigates the limitations of training the GAN framework; it enables training on a limited number of labeled data, prevents overfitting to a specific data domain source, prevents mode collapse, and enables multi-class image classification.

\section{MAVEN Architecture}

\begin{figure}
\centering
\includegraphics[width=\linewidth]{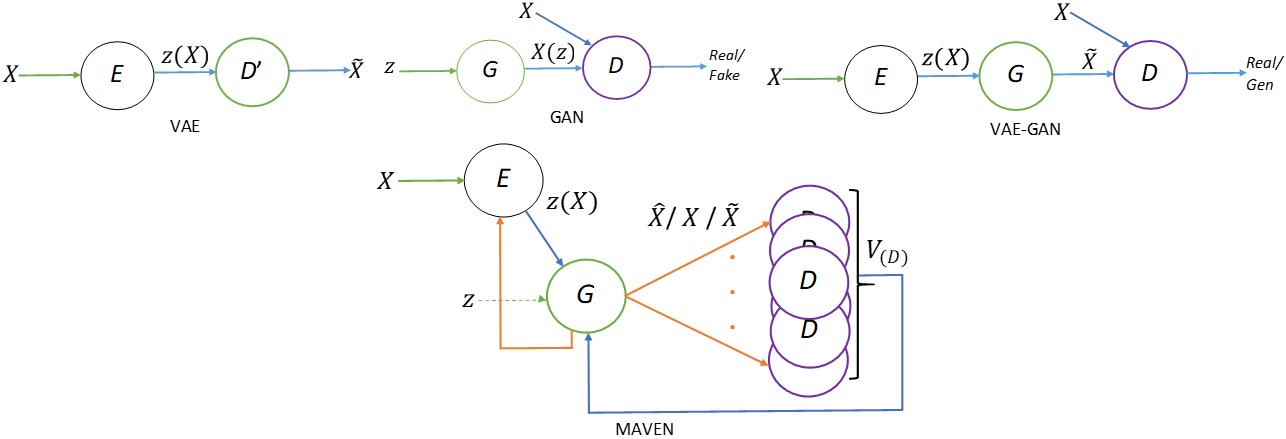}
\caption{Our MAVEN architecture compared to those of VAE, GAN, and VAE-GAN. In the MAVEN, inputs to $D$ can be real data $X$, generated data $\hat{X}$, or $\tilde{X}$. An ensemble ensures the combined feedback from the discriminators to the generator.} 
\label{fig:archs}
\end{figure}

Fig.~\ref{fig:archs} illustrates the preliminary models building up to our MAVEN architecture. 

The VAE is an explicit generative model that uses two neural nets---an
encoder $E$ and decoder $D^\prime$. Network $E$ learns an efficient compression of
the real data point $x$ into a lower dimensional latent representation
space $z(x)$; i.e., $q_\lambda(z\vert x)$. With neural network
likelihoods, computing the gradient becomes intractable. However via
differentiable, non-centered re-parameterization, sampling is performed
from an approximate function $q_{\lambda}(z\vert x) = N(z;
\mu_{\lambda}, \sigma_{\lambda}^2)$, where $z = \mu_\lambda +
\sigma_\lambda \odot \hat{\varepsilon}$ with $\hat{\varepsilon} \sim
N(0, 1)$. Encoder $E$ results in $\mu$ and $\sigma$, and with the
re-parameterization trick, $z$ is sampled from a Gaussian
distribution. Then with $D^\prime$, new samples are generated or real data
samples are reconstructed. So, $D^\prime$ provides parameters for the real
data distribution; i.e., $p_\lambda(x\vert z)$. Later, a sample drawn
from $p_\phi(x\vert z)$ may be used to reconstruct the real data by
marginalizing out $z$.

The GAN is an implicit generative model where a generator $G$ and a discriminator $D$ compete in a mini-max game over the training data to improve their performance. Generator $G$ tries to mimic the underlying distribution of the training data and generates fake samples while discriminator $D$ learns to discriminate fake generated samples from real samples. The GAN model is trained on the following objectives:
\begin{align}
\label{eqn:discriminator}
\max_{D}V(D) &= E_{x\sim p_\text{data}(x)}[\log D(x)] + E_{x\sim p_g(z)}[\log(1 - D(G(z))];\\
\label{eqn:generator}
\min_{G}V(G) &= E_{x \sim p_z(z)}[\log(1 - D(G(z))].
\end{align}
$G$ takes a noise sample $z\sim p_g(z)$ and learns to map into image space as if they are coming from the original data distribution $p_\text{data}(x)$. The discriminator $D$ takes either real image data or fake image data as the input and provides feedback to the generator $G$, regarding whether the input to $D$ is real or fake. $D$ wants to maximize the likelihood for real samples and minimize the likelihood of generated samples. On the other hand, $G$ wants $D$ to maximize the likelihood of generated samples. A Nash equilibrium state is possible when $D$ can no longer distinguish real and generated samples meaning that the model distribution will be the same as the data distribution.  

\begin{figure}
    \centering
    \includegraphics[width=0.75\linewidth]{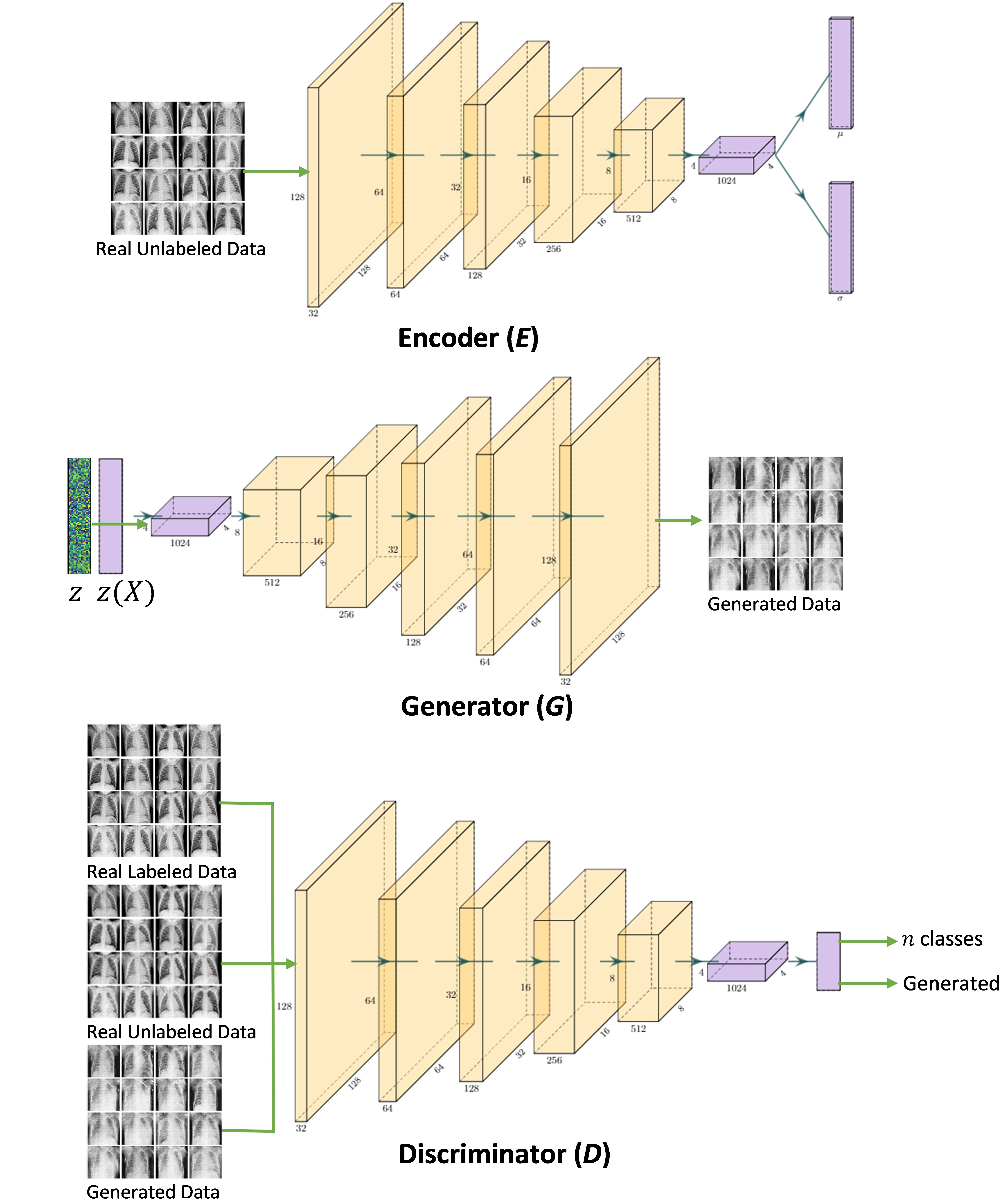}
    \caption{The three convolutional neural networks, $E$, $G$, and $D$, in the MAVEN.}
    \label{fig:arch_details}
\end{figure}

\citet{makhzani2015adversarial} proposed the adversarial training of VAEs; i.e., VAE-GANs. Although they kept both $D^\prime$ and $G$, one can merge $D^\prime$ and $G$ since both can generate data samples from the noise samples of the representation $z$. In this case, $D$ either receives generated samples $\tilde{x}$ via $G$ or fake samples $\hat{x}$, and real data samples $x$. Although $G$ and $D$ compete against each other, at some point the feedback from $D$ becomes predictable for $G$ and it keeps generating samples from the same class. At that time, the generated samples lack variety. Fig.~\ref{fig:gan_collapse}c shows an example where all the generated images are of the same class. \citet{durugkar2016generative} proposed that using multiple discriminators in a GAN model helps improve performance, especially resolving the mode collapse issue. Moreover, a dynamic ensemble of multiple discriminators has recently been proposed, addressing the same issue \citep{mordido2018dropout}. 

In our MAVEN, the VAE-GAN combination is extended to have multiple discriminators aggregated in an ensemble layer. As in a VAE-GAN, the MAVEN has three components $E$, $G$, and $D$; all are convolutional neural networks with convolutional or transposed convolutional layers (Fig.~\ref{fig:arch_details}). $E$ takes real samples and generates a dimensionality-reduced representation $z(x)$. $G$ can take samples from noise distribution $z\sim p_g(z)$ or sampled noise $z(x)\sim q_\lambda(x)$, and it generates fake or completely new samples. $D$ takes inputs from distributions of real labeled data, real unlabeled data, and fake generated data. Fractionally strided convolutions are performed in $G$ to obtain the image dimension from the latent code.
The goal of an autoencoder is to maximize the Evidence Lower Bound (ELBO).  The intuition here is to show the network more real data. The more real data that it sees, the more evidence is available to it and, as a result, the ELBO can be maximized faster.  $K$ discriminators are collected in an ensemble layer and the combined feedback 
\begin{equation}
\label{eqn: mean_ensemble}
    V(D) = \frac{1}{K}\sum_{i=1}^K w_iD_i
\end{equation}
is passed to $G$. In order to randomize the feedback from multiple discriminators, a single discriminator is randomly selected.

\begin{algorithm}[t]
\caption{MAVEN Training procedure.\\ 
$m$ is the number of samples; $B$ is the minibatch-size; and $K$ is the number of discriminators.}
 \label{alg:MAVEN}
\begin{algorithmic}
\STATE $steps \leftarrow \frac{m}{B}$
\FOR{each {\bf epoch}}
\FOR{each step in $steps$}
\FOR{k = 1 to {\bf K}}
\STATE Sample minibatch $z_i;{z^{(1)},\dots,z^{(m)}},z_i\sim p_g(z)$
\STATE Sample minibatch $x_i; {x^{(1)},\dots,x^{(m)}}, x_i\sim p_\text{data}(x)$
\STATE Update discriminator $D_k$ by ascending along its gradient:
\begin{equation*} \nabla_{\theta_{D_k}} \frac{1}{m}\sum_{i=1}^m[\log D_k(x_i) + \log(1 - D_k(G(z_i)))]
\end{equation*}
\ENDFOR
\STATE Sample minibatch $z_{k_i}, i=1,\dots, m, k=1,\dots, K, z_{k_i}\sim p_g(z)$

\IF{ensemble is `mean'}
\STATE Assign weights $w_k$ for each of the discriminators $D_k$
\STATE Determine the mean discriminator $D_\mu$ of the discriminators $D_1,...,D_k$
\begin{equation*}
    D_\mu = \frac{1}{K}\sum_i^K w_iD_i 
\end{equation*}
\ENDIF

\STATE Update the generator $G$ by descending along its gradient from the ensemble of discriminator $D_{\mu}$:
\begin{equation*} \nabla_{\theta_{G}} \frac{1}{m}\sum_{i=1}^m[\log(1 - D_{\mu}(G(z_i)))]
\end{equation*}
\STATE Sample minibatch $x_i; {x^{(1)},\dots,x^{(m)}}, x_i\sim p_\text{data}(x)$
\STATE Update encoder along its expectation function:
\begin{equation*}
\nabla_{\theta_{ E_{q_\lambda(z|x)}}} \left[\log \frac{p(z)}{q_\lambda (z|x)} \right]
\end{equation*}
\ENDFOR
\ENDFOR
\end{algorithmic}
\end{algorithm}

\section{Semi-Supervised Learning}

The overall training procedure of the proposed MAVEN model is presented in Algorithm~\ref{alg:MAVEN}. In the forward pass, the real samples to $E$ and noise samples to $G$ are presented multiple times for the presence of multiple discriminators. In the backward pass, the combined feedback from the $D$s is determined and passed to $G$ and $E$.

In the original image generator GAN, $D$ works as a binary classifier---it classifies the input image as real or synthetic. In order to facilitate the training for a $n$-class classifier, the role of $D$ is changed to an $(n+1)$-classifier. For multiple logit generation, the sigmoid function is replaced by a softmax function. Now, it can receive an image $x$ as input and outputs an $(n+1)$-dimensional vector of logits $\{{l}_1, {l}_2,\dots,{l}_{n+1}\}$. These logits are finally transformed into class probabilities for the final classification. Class ${(n+1)}$ is for the fake data and the remaining $n$ are for the multiple labels in the real data. The probability of $x$ being fake is
\begin{equation}
\label{eqn:fake_prob}  
    p(y = n+1 | x) = \frac{\exp(l_{n+1})}{\sum_{j=1}^{n+1}\exp(l_j)},
\end{equation}
and the probability that $x$ is real and belongs to class $i$ is
\begin{equation}
\label{eqn:real_prob}  
    p(y= i|x, i< n+1) = \frac{\exp(l_i)}{\sum_{j=1}^{n+1}\exp(l_j)}.
\end{equation}
As a semi-supervised classifier, the model only takes labels for a small portion of training data. For the labeled data, it is then like supervised learning, while it learns in an unsupervised manner for the unlabeled data. The advantage comes from generating new samples. The model learns the classifier by generating samples from different classes.  

\subsection{Losses}

Three networks $E$, $G$, and $D$ are trained on different objectives. $E$ is trained on maximizing the ELBO, $G$ is trained on generating realistic samples, and $D$ is trained to learn a classifier that classifies fake generated samples or particular classes for the real data samples.  

\paragraph{D Loss:} 
Since the model is trained on both labeled and unlabeled training data, the loss function of $D$ includes both supervised and unsupervised losses. When the model receives real labeled data, it is just the standard supervised learning loss
\begin{equation}
\label{eqn:supervised_loss}
    L_{D_\text{supervised}} =
    - \mathbb{E}_{\mathbf{x},y\sim p_\text{data}} \log[p(y = i|\mathbf{x}, i< n+1)].
\end{equation}
When it receives unlabeled data from three different sources, the unsupervised loss contains the original GAN loss for real and fake data from two different sources: fake1 directly from $G$ and fake2 from $E$ via $G$. The three losses 
\begin{equation}
\label{eqn:D_real}  
    L_{D_\text{real}} = - \mathbb{E}_{x \sim p_\text{data}} \log [ 1 - p(y = n+1 | \mathbf{x})],
\end{equation}  
\begin{equation}
\label{eqn:D_fake}
L_{D_\text{fake1}} = - \mathbb{E}_{\hat{x} \sim G} \log [p(y = n+1 | \hat{\mathbf{x}})],
\end{equation}
and
\begin{equation}
\label{eqn:D_recon}
    L_{D_\text{fake2}}=-\mathbb{E}_{\tilde{x} \sim G} \log [p(y = n+1 | \tilde{x})],
\end{equation}
are combined as the unsupervised loss in $D$:
\begin{equation}
\label{eqn:D_unsupervised} 
L_{D_\text{unsupervised}} = L_{D_\text{real}} + L_{D_\text{fake1}} + L_{D_\text{fake2}}.
\end{equation}

\paragraph{G Loss:} 
For $G$, the feature loss is used along with the original GAN loss. Activation $f(x)$ from an intermediate layer of $D$ is used to match the feature between real and fake samples. Feature matching has shown a lot of potential in semi-supervised learning \citep{salimans2016improvedT}. The goal of feature matching is to push the generator to generate data that matches real data statistics. The discriminator specifies those statistics; it is natural that $D$ can find the most discriminative features in real data against data generated by the model:
\begin{equation}
\label{eqn:G_feature}  
    L_{G_\text{feature}} = || \mathbb{E}_{x \sim p_\text{data}} f(x) - \mathbb{E}_{\hat{x} \sim G}f(\hat{x}) ||^2_2.
\end{equation}  
The total $G$ loss becomes the combined feature loss and $G$ costs maximizing the log-probability of $D$ making a mistake for generated data (fake1/fake2). Therefore, the $G$ loss 
\begin{equation}
\label{eqn:G_loss}  
    L_G = L_{G_\text{feature}} + L_{G_\text{fake1}} + L_{G_\text{fake2}}.
\end{equation}
is the combination of three losses, (\ref{eqn:G_feature}),
\begin{equation}
\label{eqn:G_fake}  
    L_{G_\text{fake1}} = - \mathbb{E}_{\hat{x} \sim G} \log [ 1 - p(y = n+1 | \hat{x})],
\end{equation} 
and
\begin{equation}
\label{eqn:G_recon}  
    L_{G_\text{fake2}} = - \mathbb{E}_{\tilde{\mathbf{x}} \sim G} \log [ 1 - p(y = n+1 | \tilde{x}].
\end{equation}

\paragraph{E Loss:} 
In the encoder $E$, the maximization of ELBO is equivalent to minimization of KL-divergence, allowing approximate posterior inferences. Therefore the loss function includes the KL-divergence and also a feature loss to match the features in the fake2 data with the real data distribution. The loss for the encoder is  
\begin{equation}
\label{eqn:E_loss}
 L_E = L_{E_\text{KL}} + L_{E_\text{feature}},
\end{equation}
where
\begin{equation}
    L_{E_\text{KL}} = -KL [q_{\lambda}(z|x)|| p(z)]
    = \mathbb{E}_{q_\lambda(z|x)} \left[\log \frac{p(z)}{q_\lambda (z|x)} \right] \approx \mathbb{E}_{q_\lambda(z|x)}
\end{equation}
and
\begin{equation}
    L_{E_\text{feature}} = || \mathbb{E}_{x \sim p_\text{data}} f(x) - \mathbb{E}_{\tilde{x} \sim G}f(\tilde{x}) ||^2_2.
\end{equation}

\section{Experiments and Results}

\subsection{Data}

We used three datasets to evaluate our MAVEN model for image generation and automatic image classification from 2D images in a semi-supervised learning scheme, and we constrained the experiments to limited labeled training data, considering that a large portion of annotation is missing; specifically:
\begin{enumerate}
\item The Street View House Numbers (SVHN) dataset \citep{netzer2011reading}. There are 73,257 digit images for training and 26,032 digit images for testing in the SVHN dataset. Out of two versions of the images, we used the version which has MNIST-like $32\times32$ pixel images centered around a single character, in RGB channels. Each of the training and test images are labeled as one of the ten digits (0--9).

\item The CIFAR-10 dataset \citep{krizhevsky2009learning}, which consists of 60,000 $32\times32$ pixel color images in 10 classes. There are 50,000 training images and 10,000 test images in the CIFAR-10 dataset. This is a 10-class classification with classes airplane, automobile, bird, cat, deer, dog, frog, horse, ship, and truck.   

\item The anterior-posterior Chest X-Ray (CXR) dataset \citep{kermany2018identifying} for the classification of pneumonia and normal images. We performed 3-class classification: normal, bacterial pneumonia, and virus pneumonia. The dataset contains 5,216 training and 624 test images.
\end{enumerate}

\subsection{Implementation Details}

To compare the image generation and multi-class classification performance of our MAVEN model, we used two baselines: DC-GAN and VAE-GAN. The same generator and discriminator architectures were used for DC-GAN and MAVEN models and the same encoder was used for the VAE-GAN and MAVEN models. For our MAVENs, we experimented with 2, 3, and 5 discriminators. In addition to using the proposed mean feedback of the multiple discriminators, we also experimented with feedback from a randomly selected discriminator. All the models were implemented in TensorFlow and run on a single Nvidia Titan GTX (12GB) GPU. For the CXR dataset, the images were normalized and resized to $128\times128$ pixels before passing them to the models, while for the SVHN and CIFAR-10 datasets, the normalized images were passed to the models in their original $(32\times32\times3)$ pixel sizes. For the discriminator, after every convolutional layer, a dropout layer was added with a dropout rate of 0.4. For all the models, we consistently used the Adam optimizer with a learning rate of $2e-4$ for $G$ and $D$, and $1e-5$ for $E$ with a momentum of 0.5. All the convolutional layers were followed by batch normalizations. Leaky ReLU activations were used with $\alpha = 0.2$. For all the experiments, only 10\% training data were used along with the corresponding labels. The classification performance was measured with cross-validation and average scores were reported after running each model 10 times.  

\subsection{Evaluation}

\paragraph{Image Generation Performance:}

There are no perfect performance metrics for the unsupervised learning in measuring the quality of generated samples. However, to assess the quality of the generated images, we employed the widely used Fr\'echet Inception Distance (FID) \citep{heusel2017gans} and a simplified version of the Descriptive Distribution Distance (DDD) \citep{imran2017optimization}. To measure the Fr\'echet distance between two multivariate Gaussians, the generated samples and real data samples are compared through their distribution statistics:
\begin{equation}
\label{eqn:fid}
\text{FID} = ||\mu_\text{data} - \mu_\text{fake}||^2 + Tr(\Sigma_\text{data} + \Sigma_\text{fake} - 2\sqrt{\Sigma_\text{data}\Sigma_\text{fake}}).
\end{equation}
Two distribution samples are calculated from the 2048-dimensional activations of the pool3 layer of Inception-v3~\citep{salimans2016improvedT}.
DDD measures the closeness of a synthetic data distribution to a real data distribution by comparing descriptive parameters from the two distributions. We propose a simplified version based on the first four moments of the distributions, computed as the weighted sum of normalized differences of moments, as follows:
\begin{equation}
\label{eqn:ddd}
    \text{DDD} = - \sum_{i=1}^{i=4} \log{w_i}|\mu_{\text{data}_i} - \mu_{\text{fake}_i}|.
\end{equation}
The higher-order moments are weighted more, as the stability of a distribution can be better represented by them. For both FID and DDD, lower scores are better.

\paragraph{Image Classification Performance:}

To evaluate model performance in classification, we used two measures: image-level classification accuracy and class-wise F1 scoring. The F1 score is
\begin{equation}
\label{eqn:f1-score}
\text{F1} = \frac{2\times \text{precision} \times \text{recall}}{\text{precision} + \text{recall}},
\end{equation}
with
\begin{equation}
\text{precision} = \frac{\text{TP}}{\text{TP} + \text{FP}} \textrm{\qquad and \qquad} \text{recall} = \frac{\text{TP}}{\text{TP} + \text{FN}},
\end{equation}
where TP, FP, and FN are the number of true positives, false positives, and false negatives, respectively.

\subsection{Results}

\subsubsection{SVHN}

For the SVHN dataset, we trained the network on $32\times32$ pixel images. From the training set, we randomly picked 7,326 labeled images and the remaining unlabeled images were passed to the network. All the models were trained for 150 epochs and then evaluated. We generated an equal number of new images as the training set size. Fig.~\ref{fig:svhn_images} presents a qualitative comparison of the generated digit images from the DC-GAN, VAE-GAN, and ALEAN models relative to the real training images, suggesting that our MAVEN-generated images are more realistic.

This was further confirmed by the FID and DDD scoring. FID and DDD measurement was performed by drawing 10,000 samples from the generated images and 10,000 samples from the real training images. The generated image quality measurement was performed for eight different models, and the resultant FID and DDD scores are reported in Table~\ref{table:fid-ddd}. For FID score calculation, the FID score is reported after running the pre-trained Inception-v3 network for 20 epochs for each model. Per the scores, the MAVEN-rand model with 3 discriminators achieved the best FID and the best DDD was achieved for the MAVEN-mean model with 5 discriminators. 

For the semi-supervised classification, both image-level accuracy and class-wise F1 scores were calculated. Table~\ref{table:svhn-accuracy} compares the classification performance of all the models for the SVHN dataset. The MAVEN model consistently outperformed the DC-GAN and VAE-GAN classifiers both in classification accuracy and class-wise F1 scores. Among all the models, our MAVEN-mean model with 2 and 3 discriminators were found to be the most accurate.   

\begin{table}[t]
\setlength{\tabcolsep}{4pt}
\centering
\caption{Minimum FID and DDD scores achieved by the DC-GAN, VAE-GAN, and MAVEN models for the CIFAR-10, SVHN, and CXR datasets.}
\label{table:fid-ddd}
\medskip
\resizebox{\linewidth}{!}{
\begin{tabular}{|c|c|c?c|c|c?c|c|c|}
            \hline
             \multicolumn{3}{|c?}{CIFAR-10}
             & 
             \multicolumn{3}{|c?}{SVHN}
             &
             \multicolumn{3}{|c|}{CXR}
             \\
             \hline
            Model &
            FID &
            DDD &
            Model &
            FID &
            DDD &
            Model &
            FID &
            DDD\\
            \cline{1-3}
            \hline
            DC-GAN &
            61.293$\pm$0.209 & 0.265 
            &
            DC-GAN &
            16.789$\pm$0.303 & 0.343
            &
            DC-GAN &
            152.511$\pm$0.370 & 0.145 
            \\
            VAE-GAN & 
            15.511$\pm$0.125 & 0.224
            &
            VAE-GAN & 
            13.252$\pm$0.001 & 0.329
            &
            VAE-GAN & 
            141.422$\pm$0.580 & 0.107
            \\
            MAVEN-mean2D & 12.743$\pm$0.242 & 0.223
            &
            MAVEN-mean2D & 11.675$\pm$0.001 & 0.309
            &
            MAVEN-mean2D & 141.339$\pm$0.420 & 0.138\\
            MAVEN-mean3D & \textbf{11.316$\pm$0.808} & \textbf{0.190}
            &
            MAVEN-mean3D & 11.515$\pm$0.065 &  0.300
            &
            MAVEN-mean3D & \textbf{140.865$\pm$0.983} &\textbf{0.018}\\
            MAVEN-mean5D & 12.123$\pm$0.140 & 0.207
            &
            MAVEN-mean5D & 10.909$\pm$0.001 & \textbf{0.294}
            &
            MAVEN-mean5D & 147.316$\pm$1.169 & 0.100\\
            MAVEN-rand2D &
            12.820$\pm$0.584&
            0.194
            & 
            MAVEN-rand2D &
            11.384$\pm$0.001 & 0.316
            &
            MAVEN-rand2D &
            154.501$\pm$0.345 & 0.038\\
            MAVEN-rand3D & 
            12.620$\pm$0.001 &
            0.202
            &
            MAVEN-rand3D & 
            \textbf{10.791$\pm$0.029} & 0.357
            &
            MAVEN-rand3D & 
            158.749$\pm$0.297 & 0.179\\
            MAVEN-rand5D & 18.509$\pm$0.001 &
            0.215
            &
            MAVEN-rand5D & 11.052$\pm$0.751 & 0.323
            &
            MAVEN-rand5D & 152.778$\pm$1.254 & 0.180\\
            \cline{4-9} 
            Dropout-GAN\citep{mordido2018dropout} & 88.60$\pm$ 0.08 & - 
            \\
           TTUR\citep{heusel2017gans} & 36.9 & -
           \\ 
           Coulomb GANs\citep{unterthiner2017coulomb} & 27.300 &- \\
            AIQN\citep{ostrovski2018autoregressive} & 49.500 & -
            \\
            SN-GAN\citep{miyato2018spectral} & 21.700 & - 
            \\
            Learned Moments\citep{ravuri2018learning} & 18.9 & - 
            \\
            \cline{1-3}
        \end{tabular}
}
\end{table}

\begin{table}[t]
\setlength{\tabcolsep}{4pt}
\centering
\caption{Average cross-validation accuracy and class-wise F1 scores for the semi-supervised classification performance comparison of the DC-GAN, VAE-GAN, and MAVEN models using the SVHN dataset.}
\label{table:svhn-accuracy}
\medskip
\resizebox{\linewidth}{!}{
\begin{tabular}{|c|c|c|c|c|c|c|c|c|c|c|c|}
            \cline{1-12} 
            Model &
            Accuracy &
            \multicolumn{10}{|c|}{F1 scores}\\
            \cline{3-12}
            &&
            0 & 
            1 & 
            2 &
            3 &
            4 &
            5 &
            6 &
            7 &
            8 &
            9 \\
            \hline
            DC-GAN & 
            0.876 &
            0.860 & 
            0.920 &
            0.890 &
            0.840 &
            0.890 & 
            0.870 &
            0.830 &
            0.890 &
            0.820 &
            0.840 \\ 
            VAE-GAN &
            0.901 &
            0.900 & 
            0.940 &
            0.930 &
            0.860 &
            0.920 & 
            0.900 &
            0.860 &
            0.910 &
            0.840 &
            0.850 \\   
            MAVEN-mean2D &
            \textbf{0.909} &
            0.890 & 
            0.930 &
            0.940 &
            0.890 &
            0.930 &
            0.900 &
            0.870 &
            0.910 &
            0.870 &
            0.890 \\       MAVEN-mean3D &
            \textbf{0.909} &
            0.910 & 
            0.940 &
            0.940 &
            0.870 &
            0.920 & 
            0.890 &
            0.870 &
            0.920 &
            0.870 &
            0.860 \\       MAVEN-mean5D &
            0.905 &
            0.910 & 
            0.930 &
            0.930 &
            0.870 &
            0.930 & 
            0.900 &
            0.860 &
            0.910 &
            0.860 &
            0.870 \\       MAVEN-rand2D &
            0.905 &
            0.910 & 
            0.930 &
            0.940 &
            0.870 &
            0.930 & 
            0.890 &
            0.860 &
            0.920 &
            0.850 &
            0.860 \\       MAVEN-rand3D &
            0.907 &
            0.890 & 
            0.910 &
            0.920 &
            0.870 &
            0.900 & 
            0.870 &
            0.860 &
            0.900 &
            0.870 &
            0.890 \\       MAVEN-rand5D &
            0.903 &
            0.910 & 
            0.930 &
            0.940 &
            0.860 &
            0.910 & 
            0.890 &
            0.870 &
            0.920 &
            0.850 &
            0.870 \\       \hline
        \end{tabular}
}
\end{table}

\begin{table}[t]
\setlength{\tabcolsep}{4pt}
\centering
\caption{Average cross-validation accuracy and class-wise F1 scores for the semi-supervised classification performance comparison of the DC-GAN, VAE-GAN, and MAVEN models using the CIFAR-10 dataset.}
\label{table:cifar10-accuracy}
\medskip
\resizebox{\linewidth}{!}{
\begin{tabular}{|c|c|c|c|c|c|c|c|c|c|c|c|}
            \cline{1-12} 
            Model &
            Accuracy &
            \multicolumn{10}{|c|}{F1 scores}\\
            \cline{3-12}
            &&
            airplane & 
            automobile & 
            bird &
            cat &
            deer &
            dog &
            frog &
            horse &
            ship &
            truck\\
            \hline
            DC-GAN & 
            0.713 &
            0.760 & 
            0.840 &
            0.560 &
            0.510 &
            0.660 & 
            0.590 &
            0.780 &
            0.780 &
            0.810 &
            0.810 \\
            
            VAE-GAN &
            0.743 &
            0.770 & 
            0.850 &
            0.640 &
            0.560 &
            0.690 & 
            0.620 &
            0.820 &
            0.770 &
            0.860 &
            0.830 \\   
            MAVEN-mean2D &
            0.761 &
            0.800 & 
            0.860 &
            0.650 &
            0.590 &
            0.750 &
            0.680 &
            0.810 &
            0.780 &
            0.850 &
            0.850 \\       
            MAVEN-mean3D &
            0.759 &
            0.770 & 
            0.860 &
            0.670 &
            0.580 &
            0.700 & 
            0.690 &
            0.800 &
            0.810 &
            0.870 &
            0.830 \\
            MAVEN-mean5D &
            \textbf{0.771} &
            0.800 & 
            0.860 &
            0.650 &
            0.610 &
            0.710 & 
            0.640 &
            0.810 &
            0.790 &
            0.880 &
            0.820 \\       MAVEN-rand2D &
            0.757 &
            0.780 & 
            0.860 &
            0.650 &
            0.530 &
            0.720 & 
            0.650 &
            0.810 &
            0.800 &
            0.870 &
            0.860 \\       MAVEN-rand3D &
            0.756 &
            0.780 & 
            0.860 &
            0.640 &
            0.580 &
            0.720 & 
            0.650 &
            0.830 &
            0.800 &
            0.870 &
            0.830 \\
            MAVEN-rand5D &
            0.762 &
            0.810 & 
            0.850 &
            0.680 &
            0.600 &
            0.720 & 
            0.660 &
            0.840 &
            0.800 &
            0.850 &
            0.820 \\       \hline
        \end{tabular}
}
\end{table}

\begin{table}
\setlength{\tabcolsep}{4pt}
\centering
\caption{Average cross-validation accuracy and class-wise F1 scores for the semi-supervised classification performance comparison of the DC-GAN, VAE-GAN, and MAVEN models using the CXR dataset.}
\label{table:chex-accuracy}
\medskip
\resizebox{0.6\linewidth}{!}{
\begin{tabular}{|c|c|c|c|c|}
            \cline{1-5} 
            Model &
            Accuracy &
            \multicolumn{3}{|c|}{F1 scores}\\
            \cline{3-5}
            &&
            Normal & 
            B-Pneumonia & 
            V-Pneumonia\\
            \hline
            DC-GAN & 
            0.461 &
            0.300 & 
            0.520 &
            0.480\\ 
            VAE-GAN &
            0.467 &
            0.220 & 
            0.640 &
            0.300\\
            MAVEN-mean2D &
            0.469 &
            0.310 & 
            0.620 &
            0.260\\
            MAVEN-mean3D &
            {\bf 0.525} &
            0.640 & 
            0.480 &
            0.480\\
            MAVEN-mean5D &
            0.477 &
            0.380 & 
            0.480 &
            0.540\\
            MAVEN-rand2D &
            0.478 &
            0.280 & 
            0.630 &
            0.310\\
            MAVEN-rand3D &
            0.506 &
            0.440 & 
            0.630 &
            0.220\\
            MAVEN-rand5D & 
            0.483 &
            0.170 &
            0.640 & 
            0.240\\
            \hline
        \end{tabular}
}
\end{table}

\subsubsection{CIFAR-10}

For the CIFAR-10 dataset, all the models were trained for 300 epochs and then evaluated. We generated an equal number of new images as the training set size.  Fig.~\ref{fig:cifar_images} visually compares the generated images from the GAN, VAE-GAN, and ALEAN models relative to the real training images.

The FID and DDD measurements were performed with the distribution of 10,000 samples drawn from the generated images and 10,000 samples from the real training images. For the FID score calculation, the pre-trained Inception-v3 network was run for 20 epochs and the FID score was recorded. The FID and DDD scores are reported in Table~\ref{table:fid-ddd}. As the tabulated results suggest, our proposed MAVEN models achieved better FID scores than some of the recently published models. Note that, those models were implemented in a different settings. As for the visual comparison, the FID and DDD scores confirmed more realistic image generation with our ALELAN models than the DC-GAN and VAE-GAN models. Except for MAVEN-mean with 2 discriminators, all other MAVEN models have smaller FID scores; MAVEN-rand with 3 discriminators has the smallest FID score among all the models. 

For the semi-supervised classification, both image-level accuracy and class-wise F1 scores were calculated. Table~\ref{table:cifar10-accuracy} compares the performance of all the models for the CIFAR-10 dataset.    

\subsubsection{CXR}

For the CXR dataset, all the models were trained for 150 epochs and then evaluated. We generated an equal number of new images as the training set size. Fig.~\ref{fig:chex_images} presents a visual comparison of synthesized and real image samples.

The FID and DDD measurements were performed for distribution of generated and real training samples, indicating that more realistic images were generated by the MAVEN models than by the GAN and VAE-GAN models. The FID and DDD scores presented in Table~\ref{table:fid-ddd} show that the mean MAVEN model with 3 discriminators (MAVEN-mean3D) has the smallest FID and DDD scores.

The classification performance reported in Table~\ref{table:chex-accuracy} suggests our proposed MAVEN model-based classifers are more accurate than the basline GAN and VAE-GAN classifiers. Among all the models, MAVEN-mean classifier with 3 discriminators found to be the most accurate in classifying the B-pneumonia and V-pneumonia from normal. However, the overall performance is not so good for the CXR dataset compared to the natural image datasets. A possible reason could be the shortage of data and the omission of a larger portion of the labels. The main issue in the medical image dataset is that, unlike natural images, every case is different than others, even though they are labeled as the same class. It may be possible to resolve this by augmenting the training set with the generated images from each of the models. However, the goal of our present work was to devise a generative model architecture that could be equally competitive as a generator and a classifier. Even with the relatively smaller dataset, the proposed MAVEN models perform better than the baseline models.  

\section{Conclusions}

We have demonstrated the advantages of an ensemble of discriminators in the adversarial learning of variational autoencoders and the application of this idea to semi-supervised classification from limited labeled data. Training our new MAVEN models on a small, labeled dataset and leveraging a large number of unlabeled examples, we have shown superior performance relative to prior GAN and VAE-GAN based classifiers, suggesting that our MAVEN models can be very effective in concurrently generating high-quality realistic images and improving multi-class classification performance. However, it remains an open problem to find the optimal number of discriminators that can perform consistently. Our future work will consider more complex image analysis tasks beyond classification and include more extensive experimentation spanning additional domains.

\appendix

\section{Comparison of Distributions}

\begin{figure}[h]
\centering
\resizebox{\linewidth}{!}{
  \begin{tabular}{@{}c|c|c@{}}
  SVHN & CIFAR-10 & CXR \\
  \hline
    \includegraphics[width=.4\textwidth, trim={1cm 0 0 0},clip]{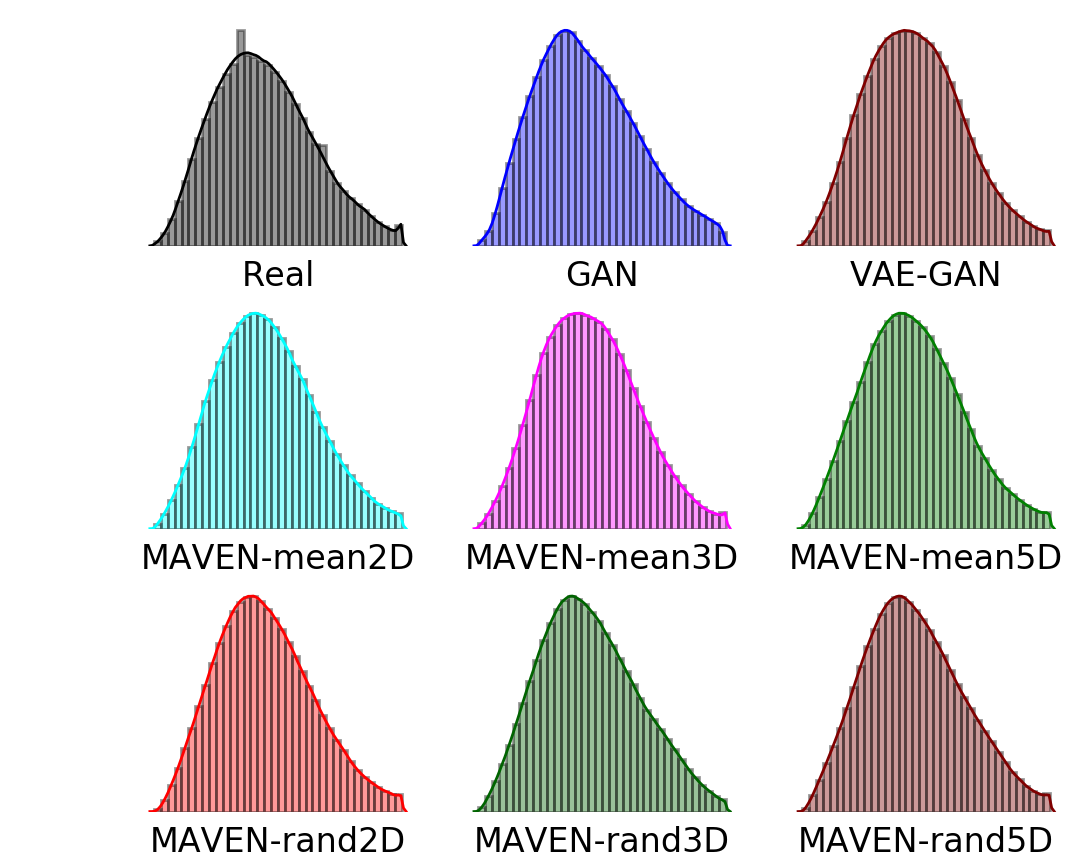} &
    \includegraphics[width=.4\textwidth, trim={0.9cm 0 0 0},clip]{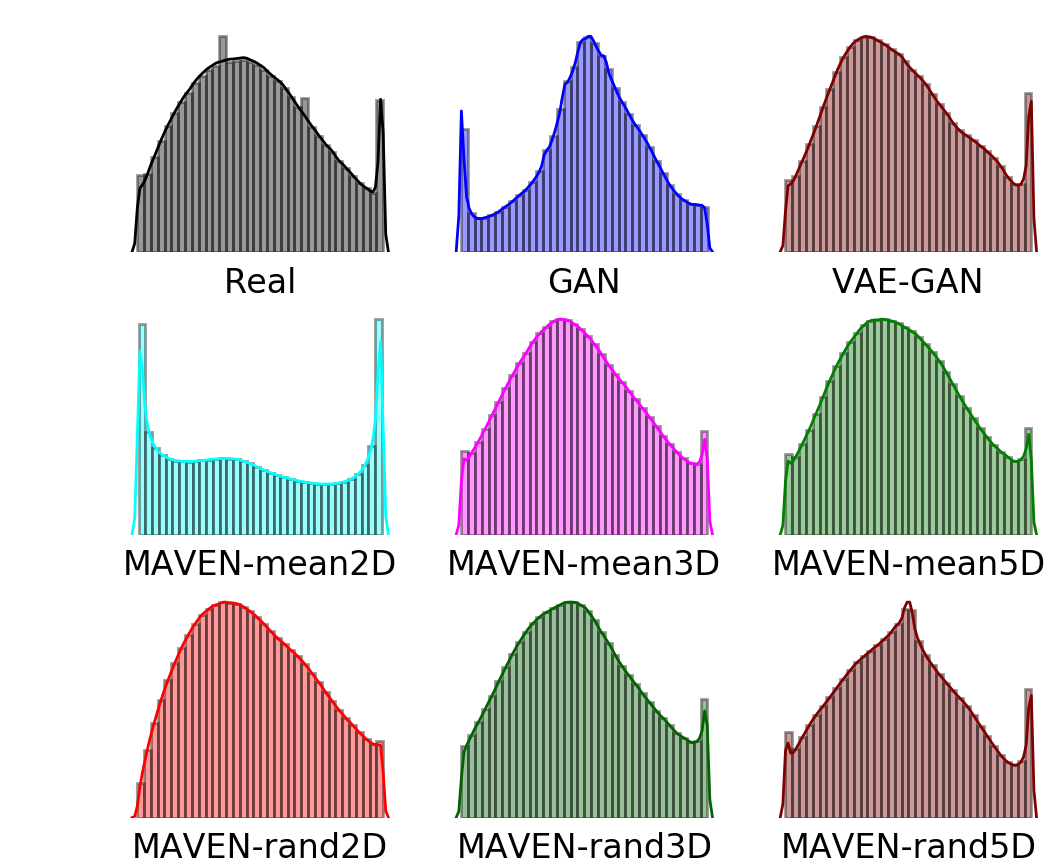} &
    \includegraphics[width=.4\textwidth, trim={1.1cm 0 0 0},clip]{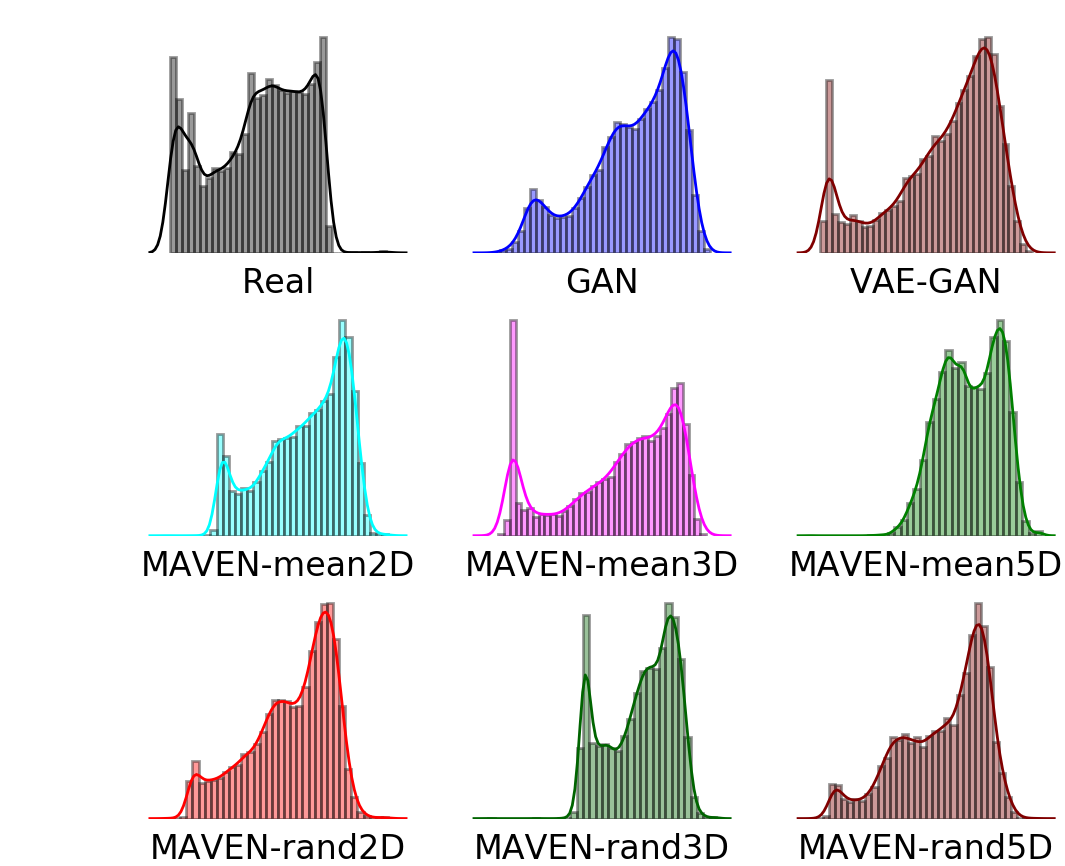}
\end{tabular}
}
  \caption{Distributions of the real training data and model-generated data from SVHN, CIFAR-10, and CXR datasets. Our MAVEN model with mean or random feedback from multiple discriminators is compared against a GAN and a VAE-GAN model.}
  \label{fig:density}
\end{figure}

Through histogram-density diagrams, Fig.~\ref{fig:density} compares the distributions of each of the models against the real distribution, showing that the distributions of images synthesized by our MAVENs are generally closer to the real image distributions for the SVHN, CIFAR-10, and CXR datasets.

\section{Comparison of Images}

\begin{figure}
\centering
\subcaptionbox{Real samples}{\includegraphics[width=0.329\linewidth]{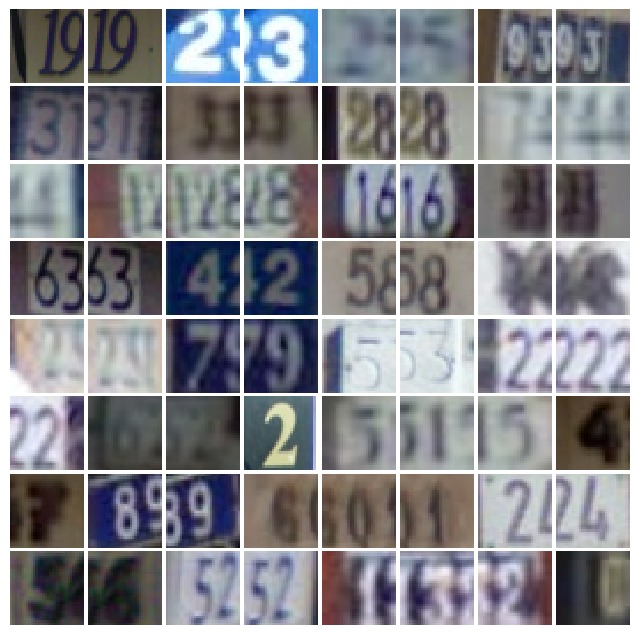}}
\hfill
\subcaptionbox{DC-GAN}{\includegraphics[width=0.329\linewidth]{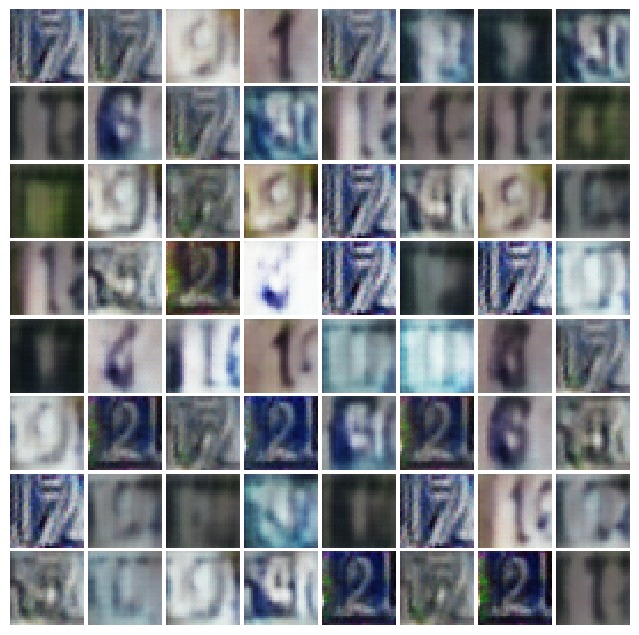}}
\hfill
\subcaptionbox{VAE-GAN}{\includegraphics[width=0.329\linewidth]{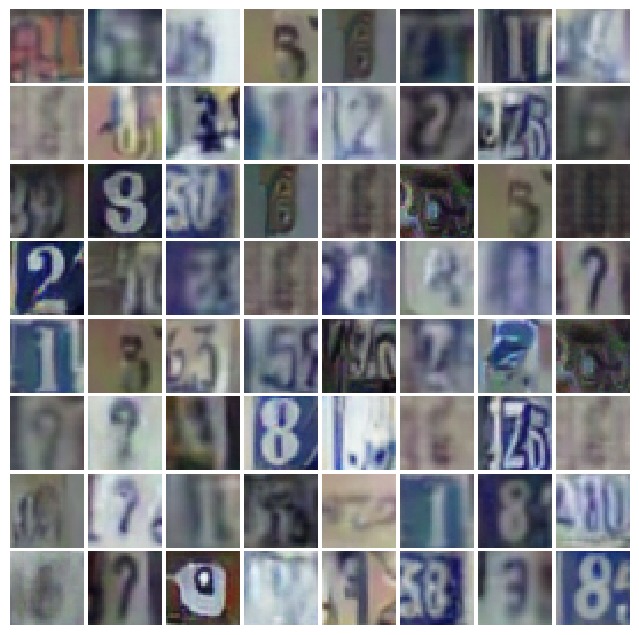}}
\\[10pt]
\subcaptionbox{MAVEN-mean2D}{\includegraphics[width=0.329\linewidth]{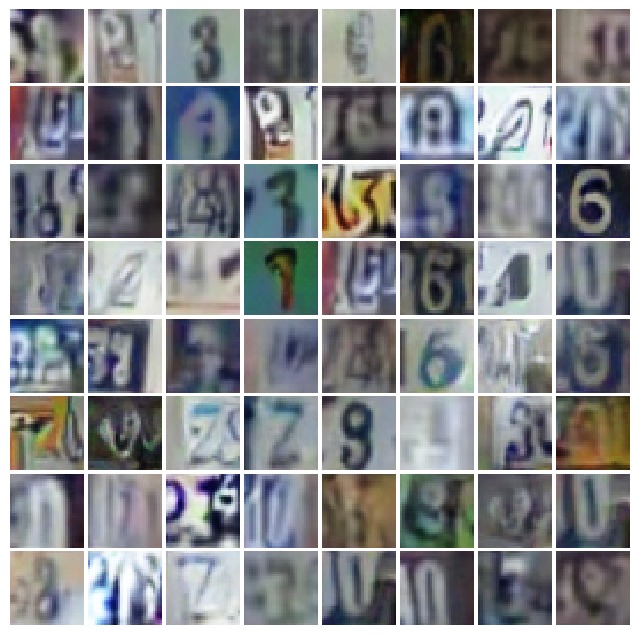}}
\hfill
\subcaptionbox{MAVEN-mean3D}{\includegraphics[width=0.329\linewidth]{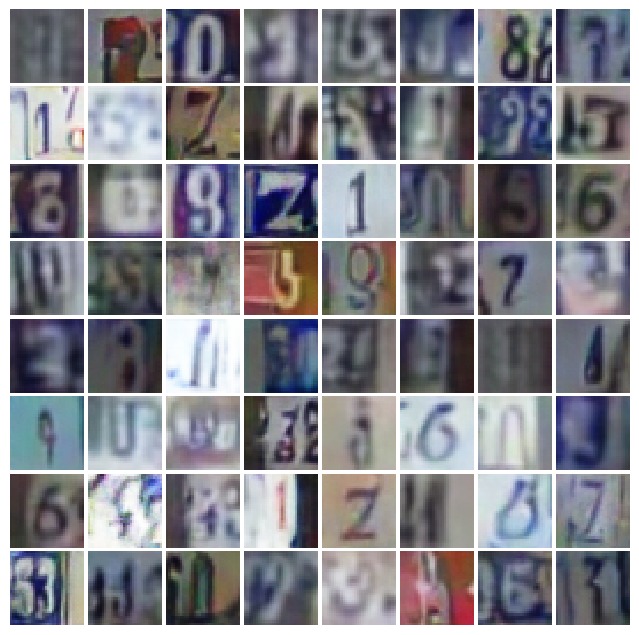}}
\hfill
\subcaptionbox{MAVEN-mean5D}{\includegraphics[width=0.329\linewidth]{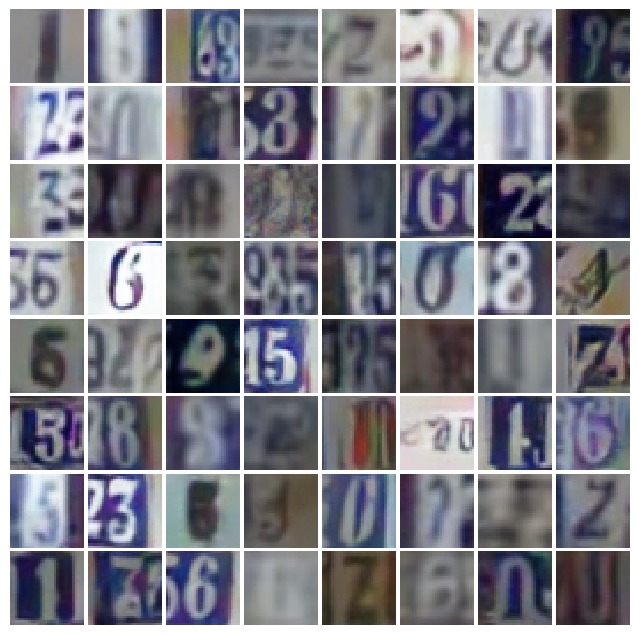}}
\\[10pt]
\subcaptionbox{MAVEN-rand2D}{\includegraphics[width=0.329\linewidth]{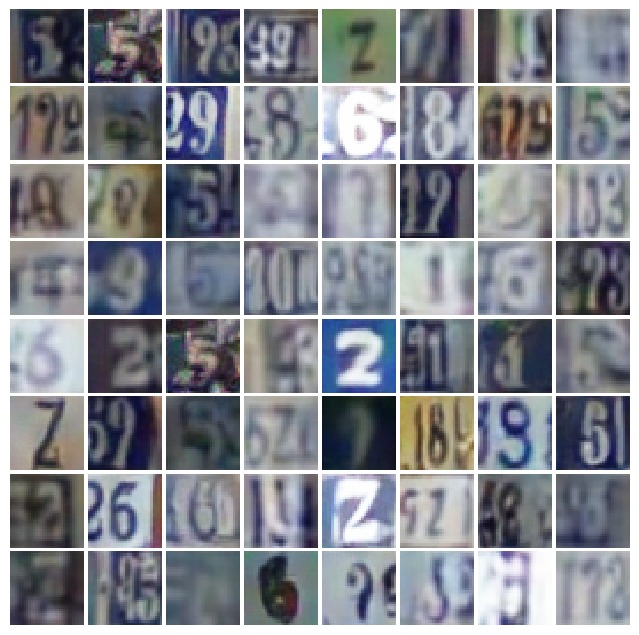}}
\hfill
\subcaptionbox{MAVEN-rand3D}{\includegraphics[width=0.329\linewidth]{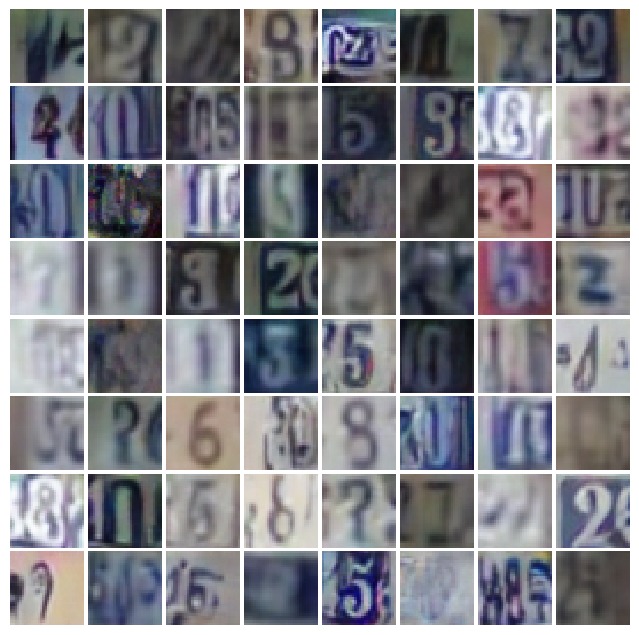}}
\hfill
\subcaptionbox{MAVEN-rand5D}{\includegraphics[width=0.329\linewidth]{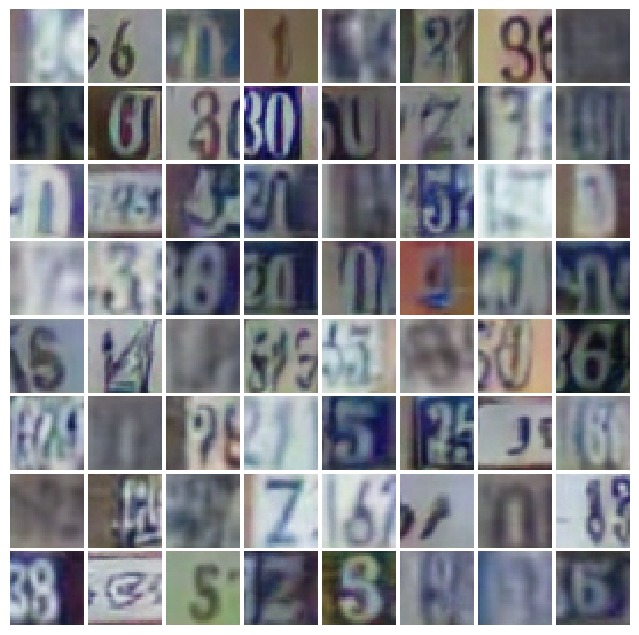}}
\caption{Visual comparison of image samples from the SVHN dataset relative to those generated by the different models.}
\label{fig:svhn_images}
\end{figure}

\begin{figure}
\centering
\subcaptionbox{Real samples}{\includegraphics[width=0.329\linewidth]{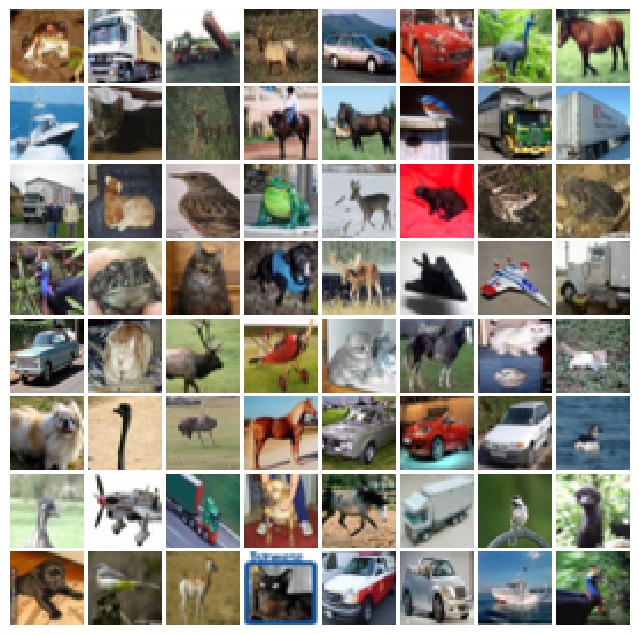}}
\hfill
\subcaptionbox{DC-GAN}{\includegraphics[width=0.329\linewidth]{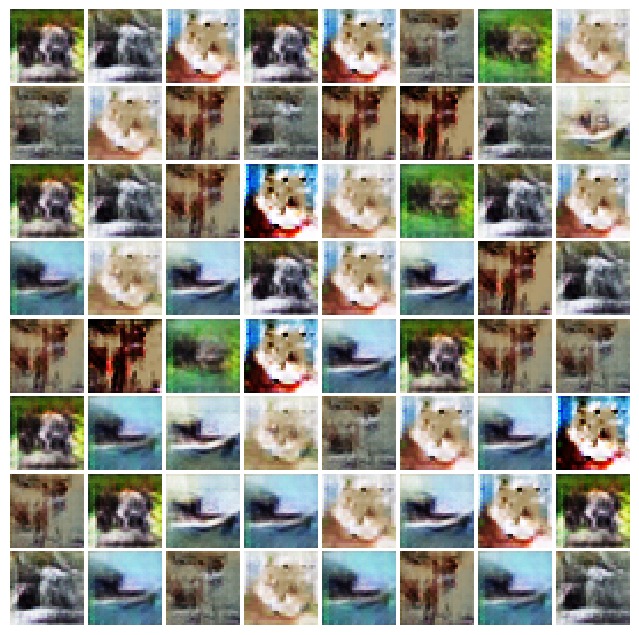}}
\hfill
\subcaptionbox{VAE-GAN}{\includegraphics[width=0.329\linewidth]{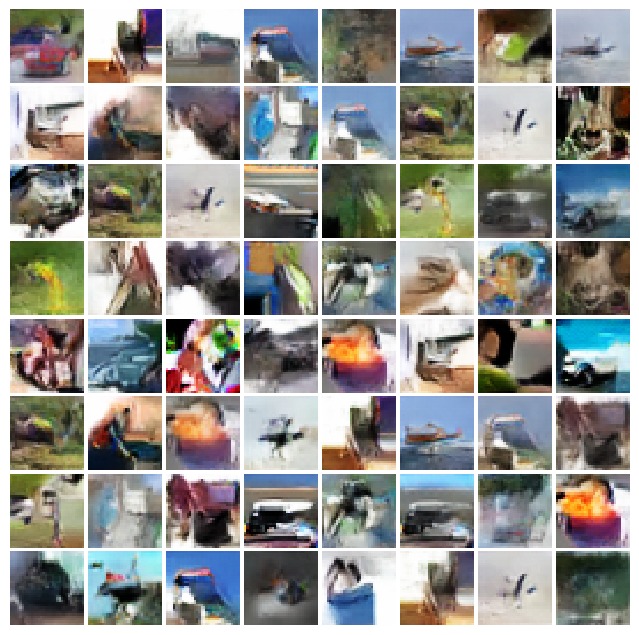}}
\\[10pt]
\subcaptionbox{MAVEN-mean2D}{\includegraphics[width=0.329\linewidth]{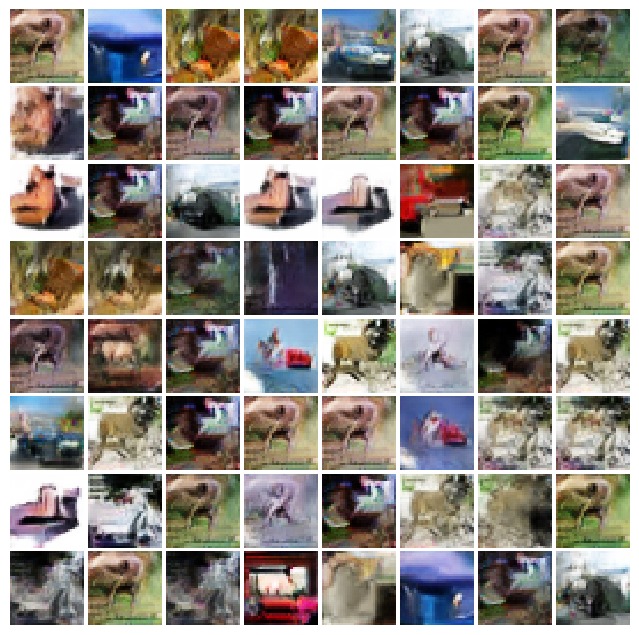}}
\hfill
\subcaptionbox{MAVEN-mean3D}{\includegraphics[width=0.329\linewidth]{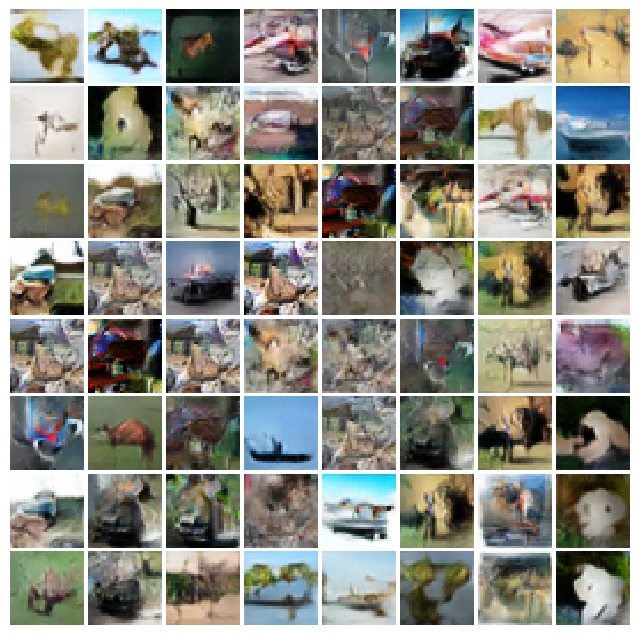}}
\hfill
\subcaptionbox{MAVEN-mean5D}{\includegraphics[width=0.329\linewidth]{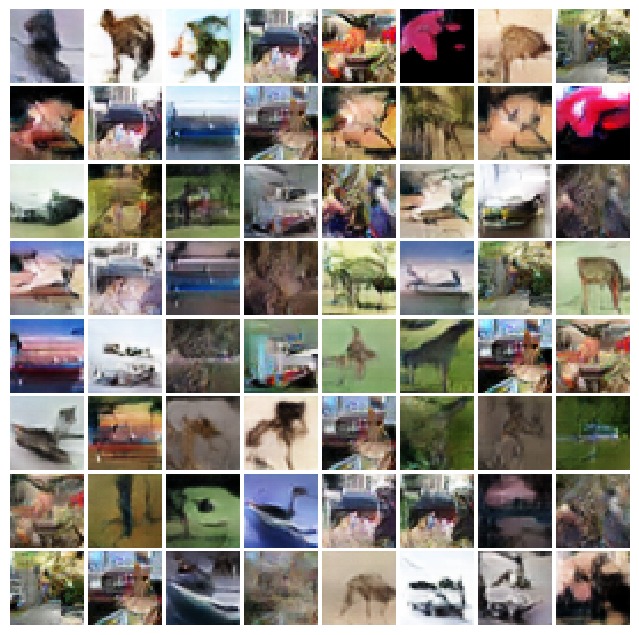}}
\\[10pt]
\subcaptionbox{MAVEN-rand2D}{\includegraphics[width=0.329\linewidth]{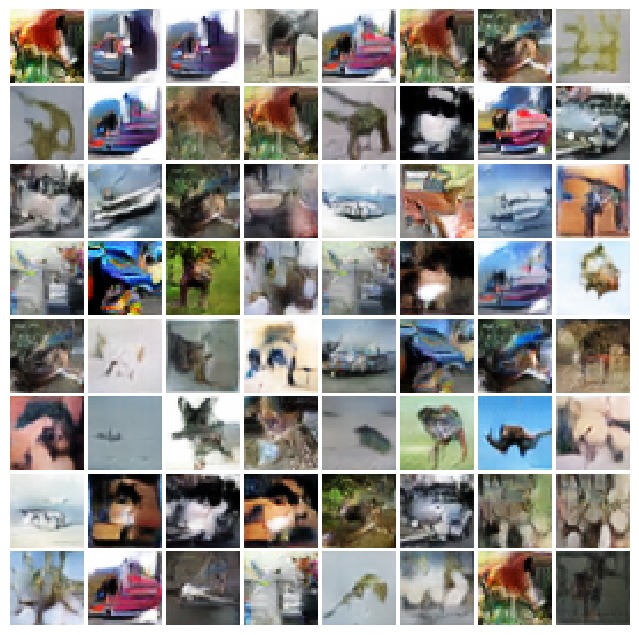}}
\hfill
\subcaptionbox{MAVEN-rand3D}{\includegraphics[width=0.329\linewidth]{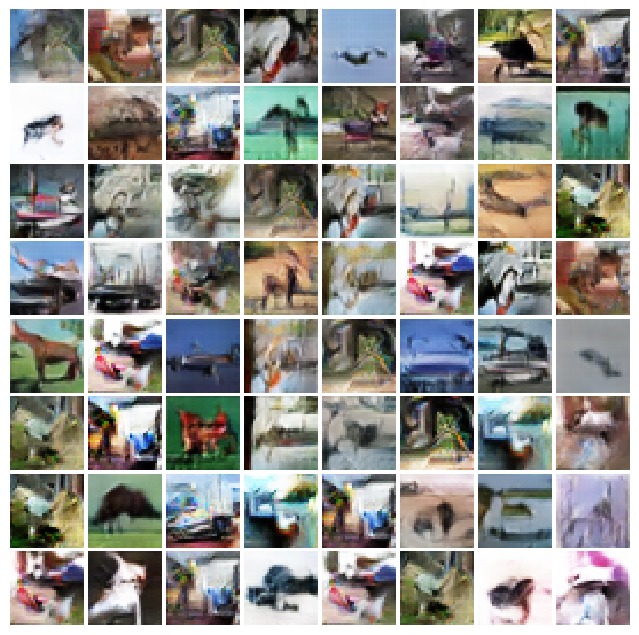}}
\hfill
\subcaptionbox{MAVEN-rand5D}{\includegraphics[width=0.329\linewidth]{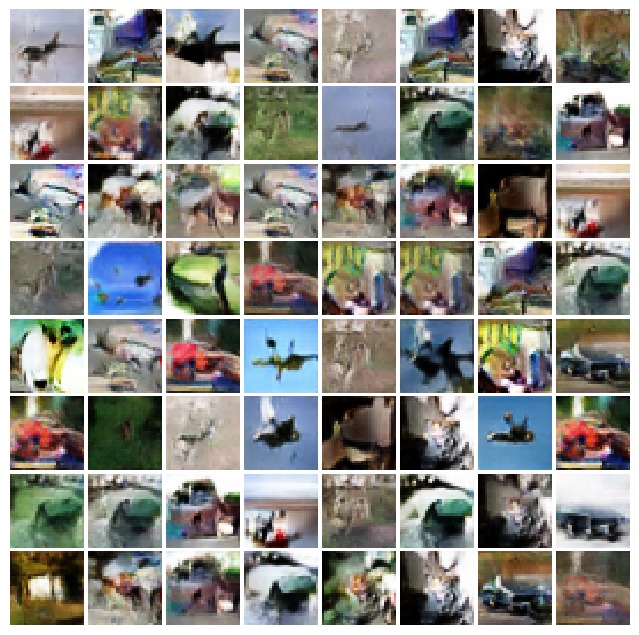}}
\caption{Visual comparison of image samples from the CIFAR-10 dataset relative to those generated by the different models.}
\label{fig:cifar_images}
\end{figure}

\begin{figure}
\centering
\subcaptionbox{Real samples}{\includegraphics[width=0.329\linewidth]{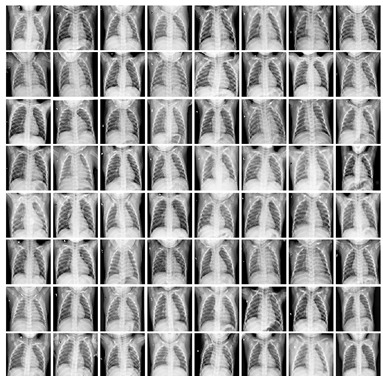}}
\hfill
\subcaptionbox{DC-GAN}{\includegraphics[width=0.329\linewidth]{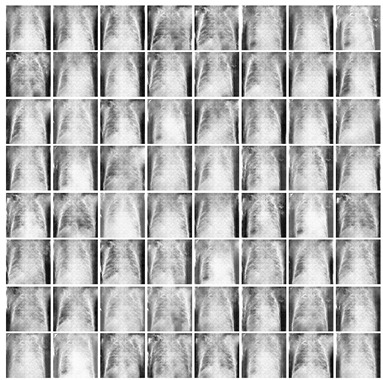}}
\hfill
\subcaptionbox{VAE-GAN}{\includegraphics[width=0.329\linewidth]{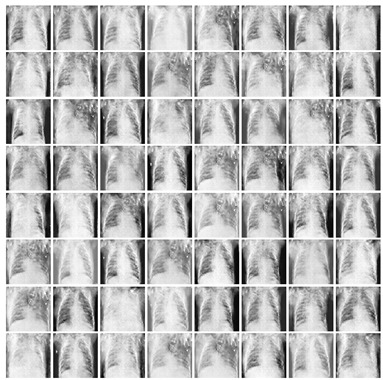}}
\\[10pt]
\subcaptionbox{MAVEN-mean2D}{\includegraphics[width=0.329\linewidth]{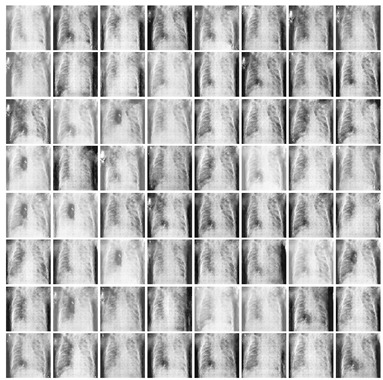}}
\hfill
\subcaptionbox{MAVEN-mean3D}{\includegraphics[width=0.329\linewidth]{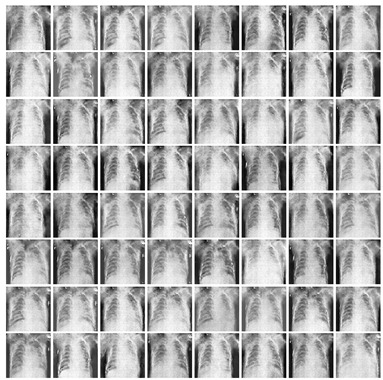}}
\hfill
\subcaptionbox{MAVEN-mean5D}{\includegraphics[width=0.329\linewidth]{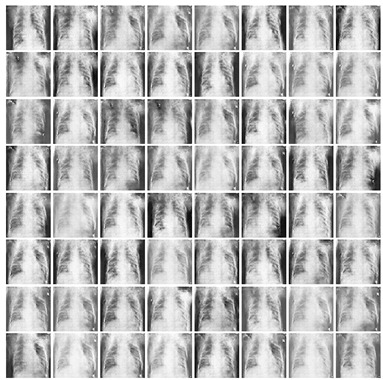}}
\\[10pt]
\subcaptionbox{MAVEN-rand2D}{\includegraphics[width=0.329\linewidth]{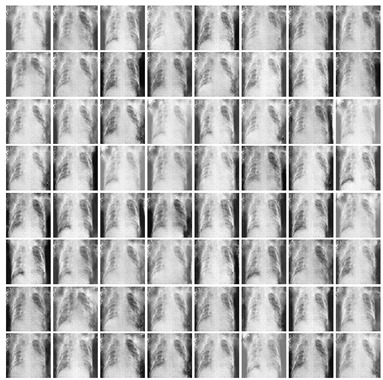}}
\hfill
\subcaptionbox{MAVEN-rand3D}{\includegraphics[width=0.329\linewidth]{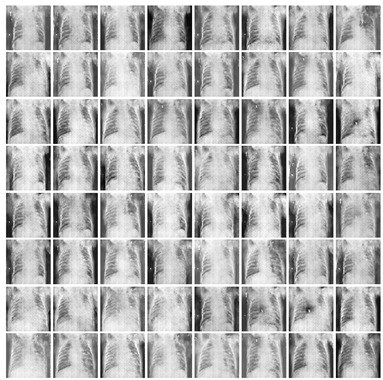}}
\hfill
\subcaptionbox{MAVEN-rand5D}{\includegraphics[width=0.329\linewidth]{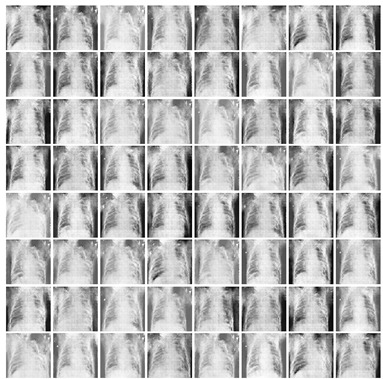}}
\caption{Visual comparison of image samples from the CXR dataset relative to those generated by the different models.}
\label{fig:chex_images}
\end{figure}

Figs.~\ref{fig:svhn_images}, \ref{fig:cifar_images}, and
\ref{fig:chex_images} present visual comparisons of image samples from
the SVHN, CIFAR-10, and CXR datasets, respectively, relative to those
generated by the different models.

\bibliographystyle{agsm}
\bibliography{references}
\addcontentsline{toc}{section}{\refname}
\end{document}